\documentclass[runningheads]{llncs}
\usepackage[T1]{fontenc}
\def\doi#1{\href{https://doi.org/\detokenize{#1}}{\url{https://doi.org/\detokenize{#1}}}}
\usepackage{graphicx}
\usepackage{listings}
\usepackage{amssymb}
\usepackage{amsmath}
\DeclareMathOperator*{\argmin}{argmin}
\usepackage{algorithm}
\usepackage{algpseudocode}
\usepackage{subcaption}
\usepackage{paralist}
\usepackage{boldline}
\usepackage{marginnote}
\usepackage[dvipsnames]{xcolor}
\definecolor{quest}{RGB}{16, 104, 16}
\newcounter{mycomments}

\begin{document}
\title{CAIPI in Practice: Towards Explainable Interactive Medical Image Classification
\thanks{Manuscript accepted at IFIP AIAI 2022.}}
\titlerunning{CAIPI in Practice}
%
\author{Emanuel Slany\inst{1} \and
Yannik Ott\inst{1} \and
Stephan Scheele\inst{1} \and
Jan Paulus\inst{2} \and \\
Ute Schmid\inst{1}}
\authorrunning{E. Slany et al.}
%
%
%
%
\institute{Fraunhofer IIS, Fraunhofer Institute for Integrated Circuits IIS, \\
Project Group Comprehensible Artificial Intelligence, \\
Bamberg, Germany \\ 
\email{\{emanuel.slany,yannik.ott,stephan.scheele,ute.schmid\}@iis.fraunhofer.de} \and
Nuremberg Institute of Technology Georg Simon Ohm, \\
Faculty of Electrical Engineering, Precision Engineering, Information Technology, \\
Nuremberg, Germany \\ 
\email{jan.paulus@th-nuernberg.de}}
\maketitle              
\begin{abstract}
  Would you trust physicians if they cannot explain their decisions to you? Medical
  diagnostics using machine learning gained enormously in importance within the last
  decade. However, without further enhancements many state-of-the-art machine learning
  methods are not suitable for medical application. The most important reasons are
  insufficient data set quality and the black-box behavior of machine learning algorithms
  such as Deep Learning models. Consequently, end-users cannot correct the model's
  decisions and the corresponding explanations. The latter is crucial for the
  trustworthiness of machine learning in the medical domain. The research field
  explainable interactive machine learning searches for methods that address both
  shortcomings. This paper extends the explainable and interactive CAIPI algorithm and
  provides an interface to simplify human-in-the-loop approaches for image
  classification. The interface enables the end-user (1) to investigate and (2) to correct
  the model's prediction and explanation, and (3) to influence the data set quality. After
  CAIPI optimization with only a single counterexample per iteration, the model achieves
  an accuracy of $97.48\%$ on the Medical MNIST and $95.02\%$ on the Fashion MNIST. This
  accuracy is approximately equal to state-of-the-art Deep Learning optimization
  procedures. Besides, CAIPI reduces the labeling effort by approximately $80\%$.
\end{abstract}
\section{Introduction}
\label{lbl:sec:intro}
Medical diagnostics based on machine learning (ML), such as Deep Learning (DL) for
visual cancer detection, have become increasingly important in the last decade~\cite{Kourou2015}.
However, on the one hand, clinicians are rarely experts in implementing ML models, on the other hand,
even if the data sets are of high quality, they are rarely intuitive for ML
engineers. Additionally, many state-of-the-art DL algorithms are black-boxes to end-users.
For ML in critical application domains like medical diagnostics, it is
crucial to close the gap between clinical domain expertise and engineering-heavy ML
methods.

This paper aims to enable domain experts such as clinicians to train and apply trustworthy ML
models. Based on the ML pipeline, starting from data preparation
to the decision-making process, we formulate three core requirements for the domain of
medical diagnostics:
\begin{inparaenum}[1)]
\item First, it is important to keep the quality of the data under control.
\item Secondly, it is necessary that the decisions of the ML model are disclosed in a
  transparent way - known as explainable machine learning (XAI) - where it is crucial in
  critical applications that models make the right decisions for the right reasons.
\item Finally, the clinical expert should be able to be involved in the optimization
  process~\cite{Holzinger2016} to allow \emph{interactive} correction of both,
  explanations and decisions. Such interactive ML methods are closely
  related to active learning \cite{Settles2012}, in which instance and label selection
  occur in the interaction between algorithm and agent.
\end{inparaenum}
By meeting these requirements, the user gains end-to-end control over the entire ML
process, involving the clinical expert \emph{interactively} in the ML process
(human-in-the-loop).

We aim to combine both, explainable and interactive machine learning, denoted by
\emph{eXplainable Interactive Machine Learning} (XIML)~\cite{Teso2019}.
Our domain of interest is the classification of medical images from diagnostics in everyday clinical
practices, such as classifying computer tomography scans into 
their corresponding categories like e.g. abdomen, chest, brain, etc. 
For our use case, we are interested in providing a visual explanation
for such a categorical classification and furthermore allow the user to correct both, the
classification and the explanation.

Our core contribution focuses on four research questions about improving the applicability
and efficiency of the CAIPI~\cite{Teso2019} algorithm, exemplary applied to the domain of medical
diagnostics. CAIPI enables model optimization to be performed while interactively
including user feedback using generated counterexamples for predictions and
explanations:
\begin{itemize}
\setlength{\itemindent}{1em}
    \item [(R1)] Do explanation corrections enhance the model's performance~\cite{Teso2019}?
    \item [(R2)] Do explanation corrections improve the explanation quality?
    \item [(R3)] Does CAIPI benefit from explanation corrections for wrong predictions?
    \item [(R4)] Is CAIPI beneficial compared to default DL optimization techniques?
\end{itemize}

We will outline the benefits of XIML for the medical domain in Section \ref{lbl:sec:related}, 
recap the most important of CAIPI's core concepts and our extensions 
in Section \ref{lbl:sec:caipi-class}, and describe our
experimental setting in Section \ref{lbl:sec:experiments}. 
We will answer the research questions in Section \ref{lbl:sec:results}.
Section \ref{lbl:sec:discussion} will discuss the results.
Finally, Section \ref{lbl:sec:conclusion} will conclude our work 
and summarize future research questions.
\section{Related Work}
\label{lbl:sec:related}
Human-in-the-loop approaches provide benefits for the application of ML in the medical
domain. Even if ML systems for medical diagnostics account for expert knowledge, they can
still suffer from a lack of trust, since knowledge bases can be manipulated
\cite{Kieseberg2015}.
The authors \cite{Kieseberg2015} propose an architecture that
allows an authorized user to enrich the knowledge base while protecting the system from
manipulation.
Although we do not provide an explicit architecture, our scope is closely
related, as we aim to enable experts to control the data quality, and to monitor and
correct the behavior of the ML system.
Apart from trustworthiness, another major benefit of interactive ML algorithms lies in
their efficiency.
For instance, extracting patient groups is more efficient when using
sub-clustering with human expert knowledge compared to traditional clustering
\cite{Hund2015}.

A central explanatory method, which CAIPI is based on, is called \emph{Local Interactive 
Model-agnostic Explanation} (LIME) \cite{Ribeiro2016}, which 
samples local interpretable features to fit a simplified and explainable surrogate model
where the surrogate model's parameters become human interpretable.
Although LIME is one of the most famous local explanation methods,
there are alternatives such as the \emph{Model Agnostic suPervised
Local Explanations} (MAPLE) method \cite{Plumb2018} that relies on linear approximation of
Random Forest models for explanations.

It is worth noting that local explanation procedures are limited to explain
single prediction instances only.
Also, they require additional explanatory models, which also
introduce uncertainty \cite{Rudin2019}.
The models do not explain the black-boxes per se, since they are ML models with
different optimization objective for themselves.
In contrast to local explanations, global explanations aim to explain the prediction model in
general. This can be achieved by approximating the complex black-box model by a simpler
interpretable model. An algorithm to approximate complex models with decision trees is
proposed by \cite{Bastani2017}. The major benefit of global explanations is that the
interpretable model mimics the explicit complex model. However, even if the resulting
models are simpler, their interpretation still requires basic ML knowledge, which apart from
the computational complexity is the major drawback for global explanations.

The authors of \cite{Schramowski2020} also extend the CAIPI algorithm specifically for DL
use cases. They introduce a loss function with additional regularization term. Large
gradients in regions with irrelevant features are penalized. Explanations for DL models
for medical image classification can also be generated with inductive logic programming
\cite{Schmid2020}. The connection of this paper with both of the previous papers appears
to be interesting for future research.
\section{Practical, Explainable, and Interactive Image Classification with CAIPI}
\label{lbl:sec:caipi-class}
In this section, we first recapitulate the mathematical foundations of LIME and
the operation of the CAIPI algorithm. Secondly, we will discuss our extension of the CAIPI
algorithm for application to complex and large image data.

The extensions will be derived by solving two problems that frequently occur with CAIPI in
practice: First, default CAIPI only receives explanation corrections, if the prediction is
correct but the explanation is wrong \cite{Teso2019}.
This seems to be inefficient, as wrong predictions are also made during the optimization.
Therefore, we extend CAIPI such that users can also correct explanations if the prediction
is wrong.
Secondly, a major contribution of our work lies in the simplification of the
human-algorithm-interaction.
In practice, optimization procedures are too complex for domain experts such as
physicians when they depend on human interaction.
To overcome this issue, we provide an universally applicable user interface.

LIME \cite{Ribeiro2016} exploits an interpretable
surrogate model to construct explanations for predictions of a complex model.
The representation of an instance is defined by $x \in \mathbb{R}^d$.
The features of an instance are transformed
into an interpretable representation $x' \in \{0,1\}^d$, where $0$ indicates the absence
and $1$ the presence of a super-pixel.  Super-pixels are contiguous patches of similar
pixels in an image.  Correspondingly, $z$ is a sample generated around the original
representation and $z'$ a sample around the interpretable representation.  The term
$\pi_x(z)$ is a proximity measure between $x$ and $z$.  The complex model is denoted by
$f(x)$ and the local explanatory model by $g \in G$, respectively, where $G$ represents
the aggregation of local explanatory models for $f(x)$.  The term $\Omega(g)$ penalizes
increasing complexity of the explanatory model.  The objective function of LIME in
\eqref{LIME_objective} aims to minimize the sum of the loss function and the penalty.

\begin{equation}
\label{LIME_objective}
    \xi(x) = \argmin_{g \in G} L(f, g, \pi_x) + \Omega(g)
\end{equation}

The locality-aware loss is defined in \eqref{LIME_loss}.  Locality-awareness means to
account for the sampling region around the representation. This is ensured by $\pi_x(z)$,
which is calculated by an exponential normalized distance function.

\begin{equation}
\label{LIME_loss}
    L(f, g, \pi_x) = \sum_{z, z' \in Z} \pi_x(z)(f(z)-g(z'))^2
\end{equation}

We make use of the Quick Shift algorithm \cite{Vedaldi2008} to partition an input image
into super-pixels and the Sparse-Linear Approximation algorithm~\cite{Ribeiro2016}
together with the loss function \eqref{LIME_loss} to generate explanations.

CAIPI \cite{Teso2019} distinguishes between a labeled data set $L$ and an unlabeled data
set $U$. It uses four components:
\begin{inparaenum}[1)]
\item The \emph{Fit} component trains a model with $L$. 
\item \emph{SelectQuery} selects a single instance from $U$. Typically, the label
belonging to this instance maximizes the loss reduction for the next optimization
step. For that, we predict the instances of $U$ and choose the instance with the lowest
prediction score.
\item \emph{Explain} applies the LIME algorithm and shows the
  prediction with its corresponding explanation.
\item Depending on the user input, the
\emph{ToCounterExamples} component generates counterexamples. The selected instance is
removed from $U$ and added to $L$ together with the generated counterexamples.
\end{inparaenum}

\begin{figure}[t]
\begin{subfigure}{.2\textwidth}
  \centering
  \includegraphics{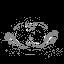}
  \caption{Original.}
  \label{aug0}
\end{subfigure}%
\begin{subfigure}{.2\textwidth}
  \centering
  \includegraphics{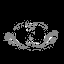}
  \caption{Features.}
  \label{aug1}
\end{subfigure}%
\begin{subfigure}{.2\textwidth}
  \centering
  \includegraphics{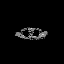}
  \caption{Scaling.}
  \label{aug2}
\end{subfigure}%
\begin{subfigure}{.2\textwidth}
  \centering
  \includegraphics{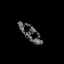}
  \caption{Rotating.}
  \label{aug3}
\end{subfigure}%
\begin{subfigure}{.2\textwidth}
  \centering
  \includegraphics{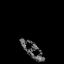}
  \caption{Translation.}
  \label{aug4}
\end{subfigure}
\caption{Data augmentation for counterexamples. Relevant features (b) are extracted from the original image (a). The features are scaled (c), rotated (d) and translated (e).}
\label{augmentation}
\end{figure}

We propose an image-specific data augmentation procedure. Fig.~\ref{augmentation} shows
decisive features of a computer tomography scan of the chest. We scale, rotate, and
translate the decisive features. The order of the augmentation is fixed. Their parameters
are random with the constraint that the resulting image must fit completely into the
original frame. The augmentation is performed with
Albumentations~\cite{Buslaev2020}. Within CAIPI, this procedure is applied to the decisive
features once when the image from $U$ is appended to $L$.

In the CAIPI optimization process in Algorithm \ref{caipi_alg}, the user provides feedback
to the most informative instance in each iteration. The model is then retrained with the
additional information. The procedure terminates when reaching a certain prediction quality of $f$
or the maximum number of iterations.

\begin{algorithm}[t]
    \centering
    \caption{CAIPI algorithm \cite{Teso2019}}\label{caipi_alg}
    \begin{algorithmic}[1]
        \Require labeled examples \(L\), unlabeled examples \(U\), iteration budget \(T\)
        \State $ f \gets Fit(L)$
        \Repeat
            \State $x \gets SelectQuery(f, U)$
            \State $\hat{y} \gets f(x)$
            \State $\hat{z} \gets Explain(f,x,\hat{y})$
            \State $\text{Present} \: x, \hat{y} \: \text{and} \: \hat{z} \: \text{to the user}$
            \State $\text{Obtain} \: y \: \text{and explanation correction} \: C$
            \State $ \{ ( \overline{x}_i, \overline{y}_i ) \}_{i=1}^c \gets ToCounterExamples(C)$
            \State $ L \gets L \cup \{ (x,y) \} \cup \{ ( \overline{x}_i, \overline{y}_i) \}_{i=1}^c$
            \State $ U \gets U \backslash ( \{x \} \cup \{ \overline{x}_i \}_{i=1}^c )$
            \State $f \gets Fit(L)$
        \Until $\text{max. number of iterations } T \text{ or min. quality of } f  \text{ is reached}$
        \State $ \textbf{return} \: f $
    \end{algorithmic}
\end{algorithm}

CAIPI distinguishes between three prediction outcome states: right for the right reasons (RRR), right for the wrong reasons (RWR), and wrong (W) \cite{Teso2019}. Whereas RRR does not require additional user input, CAIPI asks the user to correct the label for W, and to correct the explanation for RWR. RWR results in augmented counterexamples, which only contain the decisive features.

At this point, we propose the following adjustment: We require the user to provide the
correct label as well as the correct explanation for case W.
Theoretically, this adjustment makes the optimization process more efficient, as
counterexamples are generated in each iteration if either the label or the explanation or
both are wrong.

\begin{figure}
  \captionsetup[subfigure]{justification=centering}
  \hspace*{-2em}
\begin{subfigure}{.54\textwidth}
  \centering
  \includegraphics[width=1\linewidth]{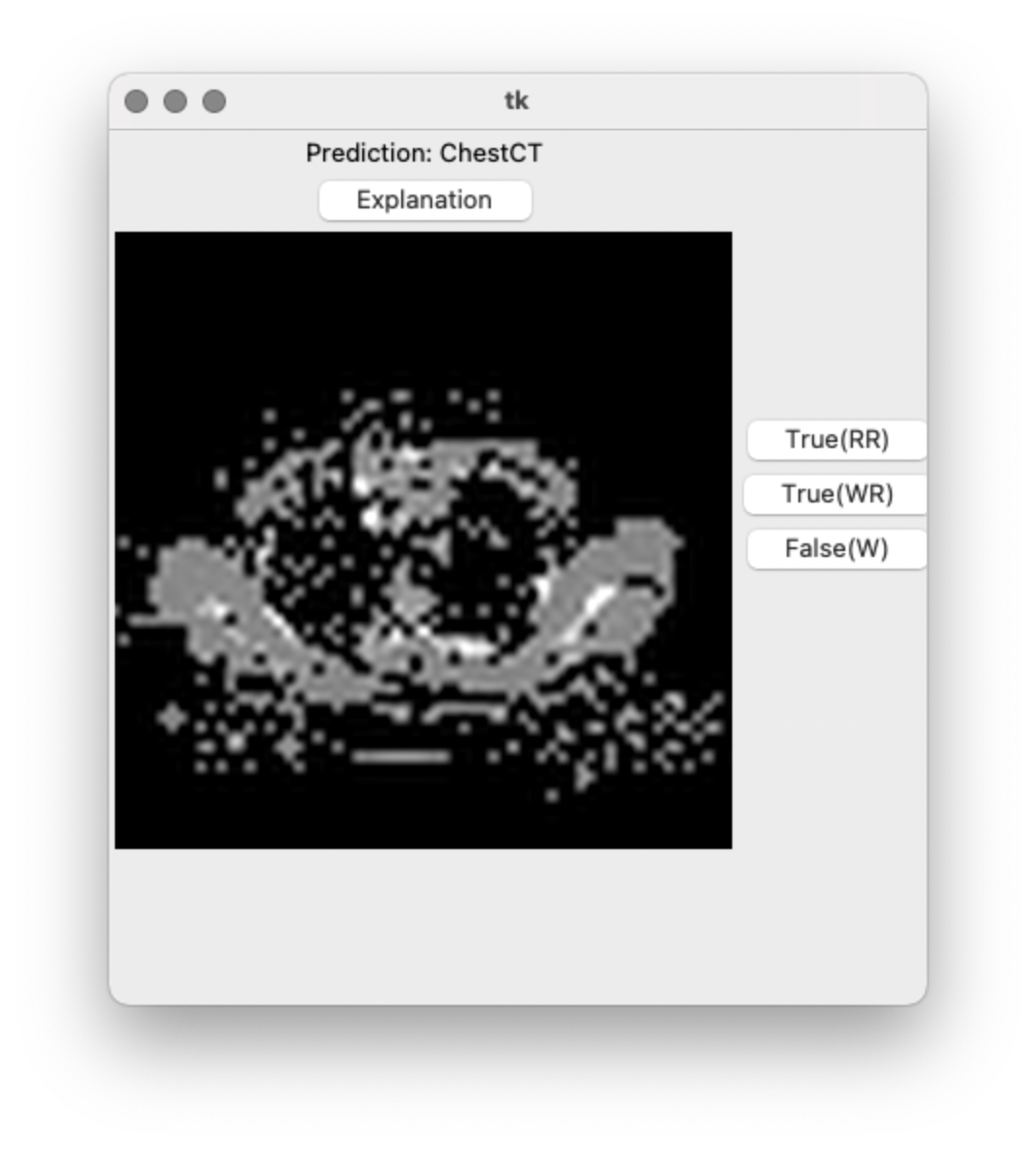}
  \caption{Prediction of the classifier.}
  \label{gui1a}
\end{subfigure}%
\quad
\begin{subfigure}{.54\textwidth}
  \centering
  \includegraphics[width=1\linewidth]{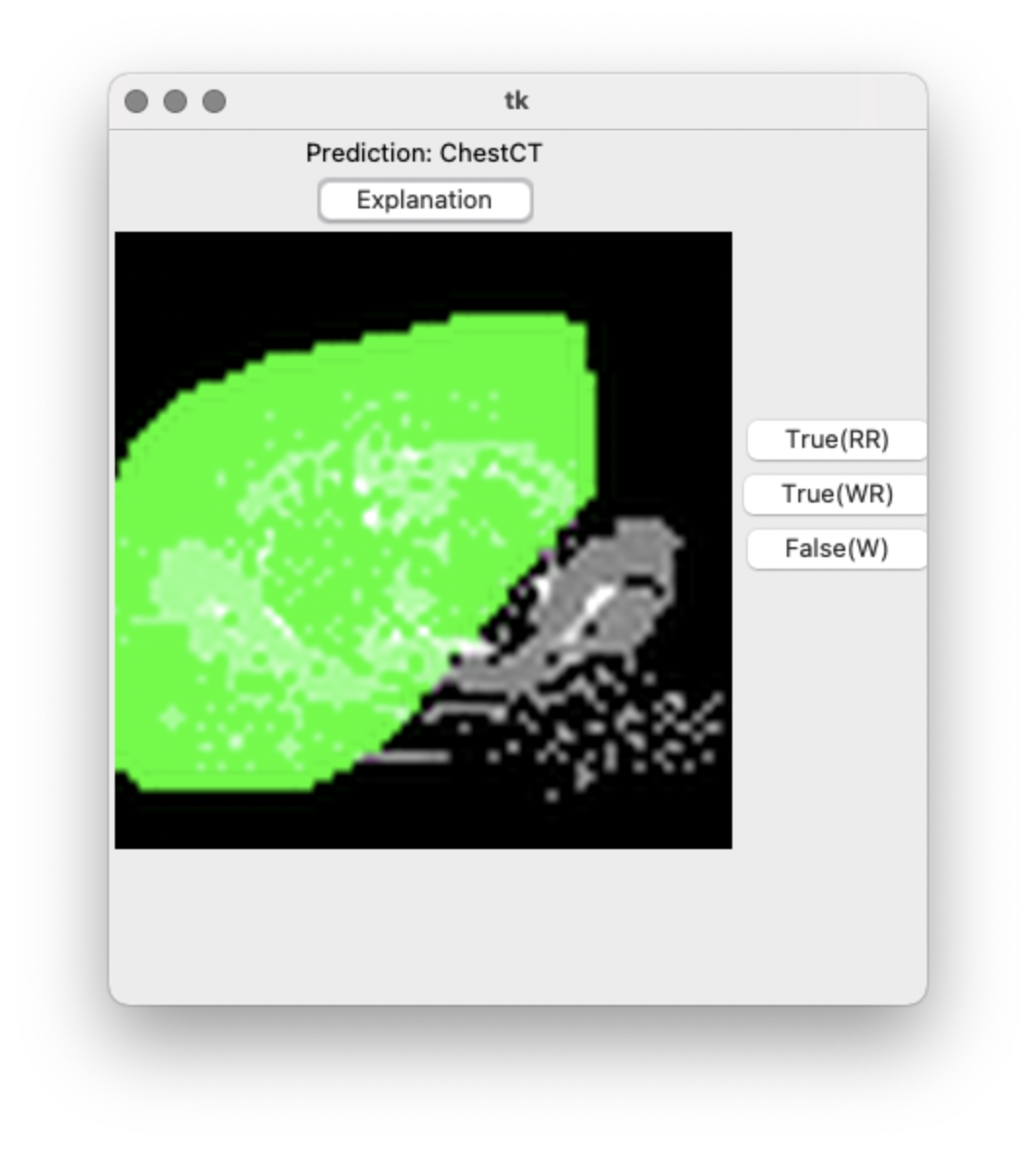}
  \caption{LIME explanation.}
  \label{gui1b}
\end{subfigure}\\
\hspace*{-2em}
\begin{subfigure}{.54\textwidth}
  \centering
  \includegraphics[width=1\linewidth]{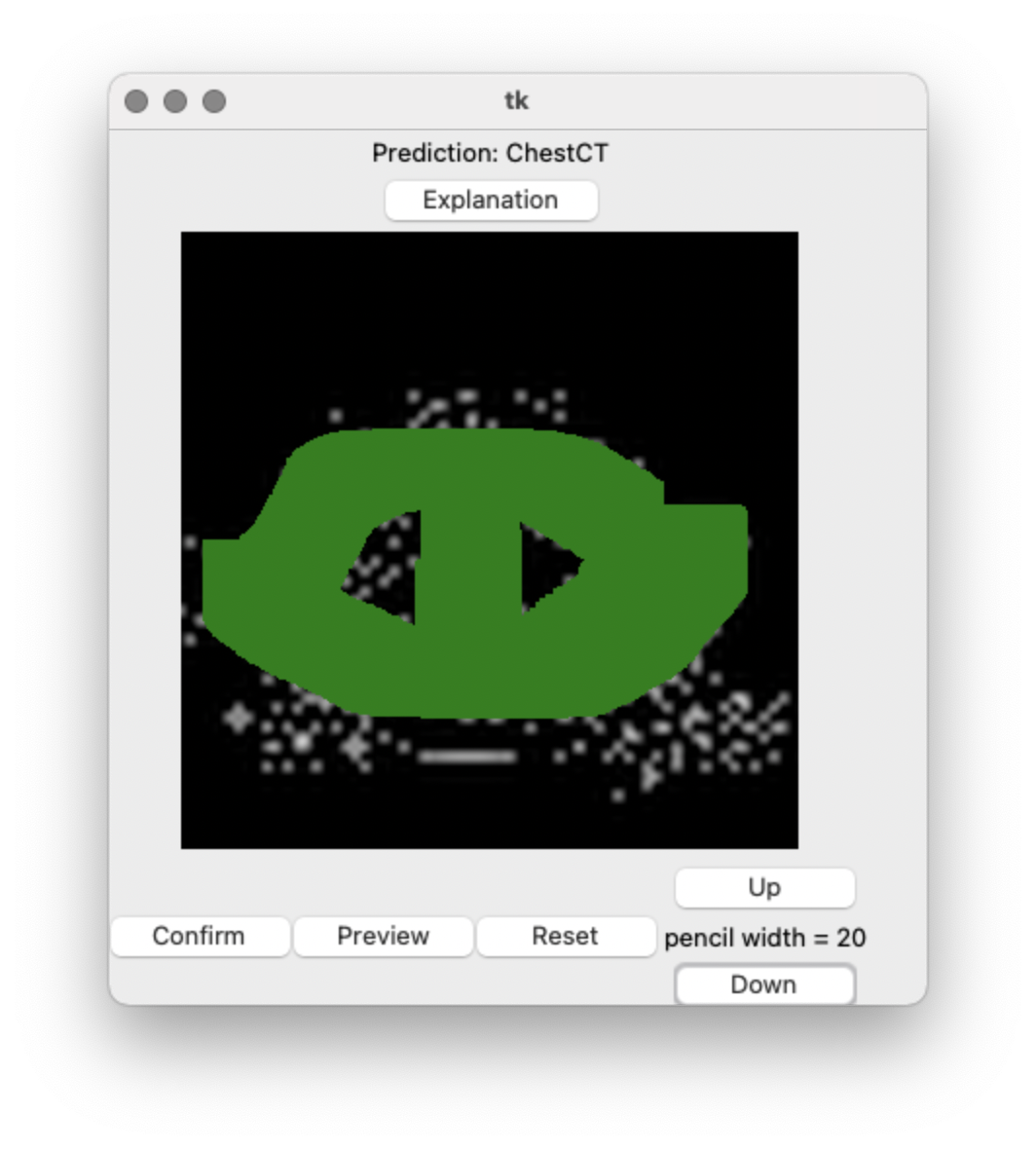}
  \caption{Correction.}
  \label{gui2a}
\end{subfigure}%
\quad
\begin{subfigure}{.54\textwidth}
  \centering
  \includegraphics[width=1\linewidth]{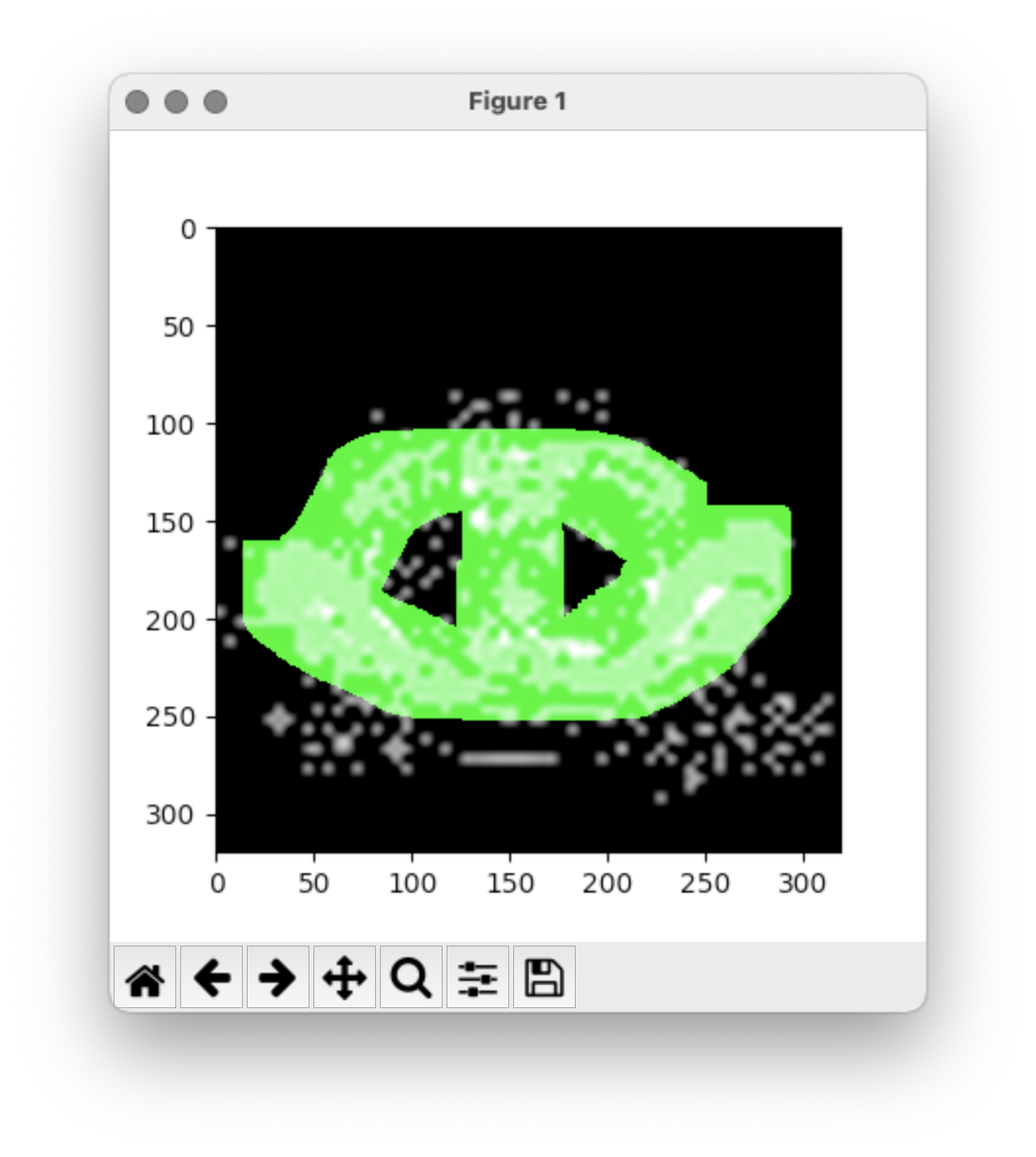}
  \caption{Corrected instance.}
  \label{gui2b}
\end{subfigure}
\caption{User interface. The prediction (a) and the explanation (b) is presented to the user. The user
can correct the model's prediction and explanation in the annotation mode (c). 
The corrected instance can be displayed (d).}
\label{gui}
\end{figure}

Fig.~\ref{gui} illustrates our proposed user interface.
The depicted example image is a computer tomography scan of the chest and
is displayed together with its prediction (Fig.~\ref{gui1a}).
Button \textit{Explanation} displays the LIME result as shown in Fig.~\ref{gui1b}.
The user can then choose whether the image was predicted correctly or not
(buttons \textit{True} or \textit{False(W)}, respectively).
In case of a correct prediction, we further distinguish between right (\textit{True(RR)}) and
wrong (\textit{True(WR)}) reasons.
This distinction maps exactly to the three cases RRR, RWR and W from CAIPI.

Fig.~\ref{gui1a} shows that the image was predicted correctly. However, as Fig.~\ref{gui1b} indicates,
the explanation is at least partly wrong, i.e.,
the instance can be considered as RWR.
The corresponding button \textit{True(WR)} opens the
annotation mode (Fig.~\ref{gui2a}), where the user can correct the
explanation. Afterwards, a newly generated explanation can be evaluated as depicted in Fig.~\ref{gui2b}.
Confirming a correction starts CAIPI's \textit{ToCounterExamples} method, which is in our
case the proposed data augmentation procedure (Fig.~\ref{augmentation}).
Note, that the same interaction applies to the W case, where the interface additionally asks for the
correct label.
For RRR, contrary, the correction mode (Fig.~\ref{gui2a} and \ref{gui2b}) 
is concealed from the user. The remaining procedure is constant with the slight modification 
that no counterexamples are generated.

The extension we propose offers great benefits for CAIPI.
First of all, CAIPI can be operated by end-users.
Secondly, it fulfills all essential requirements defined in the Introduction, Section~\ref{lbl:sec:intro}.
CAIPI shows its prediction and explanation to the end-user in each
optimization iteration, and if necessary the end-user can correct both.
Furthermore, the end-user (which typically is a domain expert) is directly responsible 
for the data set quality, as CAIPI asks to add instances to the training data set iteratively 
and the end-user can ensure correct labels and emphasize correct explanations.

\section{Experiments}
\label{lbl:sec:experiments}
For our experiments, we use two classes of the Medical MNIST data set
\cite{MNIST2021,Lozano2017} (chest and abdomen computer tomography scans) and two classes
of the Fashion MNIST data set \cite{FashionMNIST} (pullover and T-shirt/top). By this
selection, we want to emphasize a challenging binary classification task. The extension to
categorical data is left as future work due to simplicity during the evaluation process.

We use a fairly simple convolutional neural network (CNN) as DL model in all
experiments. It has a single convolutional layer with only $2$ filters, a $9$x$9$ kernel
size, and stride parameter $1$. It is followed by a pooling layer with kernel-size $8$x$8$
and stride parameter $8$. It follows a single linear layer with $98$ neurons, a dropout
layer with $0.5$ dropout-rate, and two fully-connected layers with $16$ and $2$
neurons. All training procedures use batch size $64$ and $5$ epochs. We use a binary cross
entropy loss function, the Adam optimization algorithm \cite{Kingma2015}, and a learning
rate of $0.001$. The CNN corresponds to the function $f$ in CAIPI.

Each CAIPI optimization starts with $100$ preliminary labeled instances $L_0$. The maximum number of
iterations is set to $100$. We do not specify any other stop criterion. R1 until R3 are evaluated with an
alternating number of counterexamples $c$. The number of counterexamples per iteration 
is $c = \{0, 1, 3,5\}$. We show results on a domain-related data set, the Medical MNIST, 
and on the Fashion MNIST, a well-known benchmark data set. We ensure balanced classes in both data sets.
Since CAIPI has 100 iterations, and $L_0$ has 100 instances, 
the final training data set will contain $200$ different instances. 
Depending on the user input, the training data set size will increase due to counterexamples.

We evaluate the prediction quality of our model in each optimization iteration with CAIPI with the
accuracy metric on dedicated test data sets with size $6,000$ for the Medical MNIST and size $4,200$ for
the Fashion MNIST. We also created test data sets to evaluate the explanation quality. 
Here, both test data sets have size 200. We annotated the true explanation for all instances. 
For the evaluation of the explanation quality, we use the \emph{Intersection over Unions} (IoU) metric. 
IoU lies in the interval $[0,1]$, where $0$ is a completely incorrect and $1$ a perfect explanation. 
This means we divide the intersection of the LIME explanation and the ground truth explanation 
by their union. We consider the average non-zero explanation score. 
Non-zero stands for excluding false predictions, since we cannot assume correct explanations 
for false predictions, and average means dividing by the number of instances 
with non-zero explanation score. 
We compare the prediction and explanation ability for generating counterexamples only for RWR predictions
versus generating counterexamples for predictions that are either RWR or W.

Furthermore, we conduct a benchmark test by using the identical DL setting and train a model 
with $14,000$ training instances for the Medical MNIST and evaluated on $6,000$ test samples.
Correspondingly, we trained with $9,800$ instances from the Fashion MNIST and 
tested with $4,200$ instances.
\section{Results}
\label{lbl:sec:results}
\begin{table}[t]
    \centering
    \caption{Maximum accuracy (\%) by number of counterexamples conditioned on data sets
      and modes. The mode RWR only generates counterexamples for Right predictions with
      Wrong Reasons, whereas the RWR $+$ W mode generates counterexamples additionally for
      Wrong predictions.}
    \medskip
    \setlength{\tabcolsep}{10pt}
    \renewcommand{\arraystretch}{1.2}    
    \begin{tabular}{c c c c c c}
    \hlineB{3}
      \textbf{Mode} & \textbf{Data}& \multicolumn{4}{c}{\textbf{Counterexamples $c$}}  \\
         & & 0 & 1 & 3 & 5 \\
        \hlineB{2}
        RWR & Medical MNIST & 96.02 & 95.42 & 94.51 & 96.75 \\
        RWR $+$ W & Medical MNIST & 96.83 & 97.48 & 96.92 & \textbf{97.52} \\
        \hline
        RWR & Fashion MNIST & \textbf{95.64} & 94.79 & 95.40 & 94.95 \\
        RWR $+$ W & Fashion MNIST & 94.33 & 95.02 & 94.24 & 94.10 \\
        \hlineB{3}
    \end{tabular}
    \label{tab1}
\end{table}

\begin{table}[t]
    \centering
    \caption{Maximum average non-zero explanation score (\%) by number of counterexamples
      conditioned on data sets and modes. 
      The mode RWR only generates counterexamples for Right predictions with
      Wrong Reasons, whereas the RWR $+$ W mode generates counterexamples additionally for
      Wrong predictions.}
    \medskip
    \setlength{\tabcolsep}{10pt}
    \renewcommand{\arraystretch}{1.2}    
    \begin{tabular}{c c c c c c}
      \hlineB{3}
        \textbf{Mode} & \textbf{Data} & \multicolumn{4}{c}{\textbf{Counterexamples $c$}}  \\
        & & 0 & 1 & 3 & 5 \\
        \hlineB{2}
        RWR & Medical MNIST & 41.59 & \textbf{44.74} & 40.36 & 42.87 \\
        RWR $+$ W & Medical MNIST & 39.64 & 42.37 & 41.12 & 40.47 \\
        \hline
        RWR & Fashion MNIST & 65.24 & 64.41 & 64.40 & 65.48 \\
        RWR $+$ W & Fashion MNIST & 64.43 & 65.76 & 63.10 & \textbf{66.38} \\
      \hlineB{3}
    \end{tabular}
    \label{tab2}
\end{table}

Table~\ref{tab1} clearly shows that the prediction quality of our model does not benefit
from an increasing number of counterexamples, as the maximum accuracy is approximately
stable over the runs. Also, Table~\ref{tab2} shows no clear trend towards an increasing
explanation quality for greater numbers of counterexamples. Thus, R1 and R2 can be
negated based on the experimental setting in Section 4. Similarly, for R3, the adjustment
of providing explanation corrections also for wrong predictions does not have positive
impact on either the explanation nor the prediction quality.

For state-of-the-art DL optimization, we achieve an accuracy of $94.67\%$ for the Medical
MNIST and $95.26\%$ for the Fashion MNIST. This accuracy is approximately equal to the CAIPI
results in Table~\ref{tab1}. Besides, CAIPI requires significantly less training data
($200$, plus counterexamples) than traditional optimization ($14,000$, respectively
$9,800$). With respect to R4, this is clear evidence that CAIPI influences the optimization
process positively.
\section{Discussion}
\label{lbl:sec:discussion}
R1 and R2 show no clear trend in favor of an increasing number of counterexamples, despite
\cite{Teso2019} states otherwise. Teso and Kersting \cite{Teso2019} induce decoy pixels
with colors corresponding to the different classes into their training data set and they
randomize the pixel colors in the test set. Whenever they create counterexamples, they
also randomize the decoy pixel color. We investigate R1 and R2 without prior data set
modification. Table~\ref{tab1} shows that default active learning ($c=0$) positively
influences the learning behavior to such an extent that there is hardly space for
improvement by extending it to XIML. This means that for future evaluations the use cases
must be sufficiently complex so that default active learning does not provide satisfactory
results.

We evaluate R2 by a dedicated test set containing annotated true explanations. We use IoU
to estimate the quality of the explanation in percent by dividing the area of intersection
by the area of union. The idea is, if predicted and annotated explanations are congruent
to each other, the explanation is perfect. We frequently observe perfectly negative
explanations, meaning that the predicted explanation highlights all image parts apart from
the annotated explanation. From a human perspective, this explanation is perfect, for IoU
it is a completely wrong explanation. Determining the quality of explanations is a
prominent research field and future work should evaluate additional metrics.

For R3, Table~\ref{tab1} and \ref{tab2} show that explanation corrections for RWR and W do
not differ significantly compared to explanation corrections for only RWR. We argue that
including more counterexamples (RWR $+$ W) can be a chance to build more robust data
set. The robustness from a statistical perspective was not addressed in this paper. It
must be included in future evaluations besides accounting for the priory mentioned
discussion points. Similar to earlier discussion points, R4 can be re-evaluated in more
complex use cases.

The data augmentation procedure plays also a major role. The procedure defined in
Fig.~\ref{augmentation} will create training data, which are fundamentally different to
the test data, since both data sets, Medical MNIST and Fashion MNIST, contain relatively
centralized images. The idea behind the proposed procedure is to force the model to
account for the decisive features without considering the feature position. This can be
enhanced by random transformations in every epoch compared to a single transformation when
the counterexamples are generated. We also expect improvement by including further
constraints to make the resulting counterexamples more realistically.

Finally, our main contribution is the simplification of the human-algorithm-interaction
with the introduced interface. We support this on theoretical basis, as the application of
CAIPI with our interface fulfills requirements, which we defined in the
Introduction, Section~\ref{lbl:sec:intro}. 
Furthermore, we give practical evidence via demonstration in
Section~\ref{lbl:sec:caipi-class}. From a psychological point of view, this is
insufficient. Therefore, our interface should be subject of psychological studies in
future.
\section{Conclusion and Future Work}
\label{lbl:sec:conclusion}
We extended the CAIPI algorithm by accounting additionally for explanation corrections if
the predictions are wrong. Moreover, we introduced an user interface for a human-in-the-loop
approach for image classification tasks. The interface enables the end-user (1) to
investigate and (2) to correct the model's prediction and explanation, and (3) to
influence the data set quality.

The experiments show that the predictive performance of state-of-the-art DL methods is
met, even though the required training data set size decreases. According to our findings,
the correlation between an increasing amount of counterexamples and higher predictive and
explanatory quality does not hold. The introduced extension that creates counterexamples
also for wrong predictions can help to build more robust data sets but does not increase
the predictive nor the explanatory quality. The proposed interface is a promising
extension for medical image classification tasks using CAIPI. The interface appears to be
transferable to every XIML approach exploiting local explanations. Evidently, CAIPI as
well as the proposed interface is transferable to any other image classification task.

The most obvious improvement is the generalization to categorical image data. This appears
to be a minor adjustment. It was neglected in this paper for the sake of simplicity during
evaluation of the experiments. Future research should also address wrong
explanations. This can be accomplished by connecting this paper with
\cite{Schramowski2020}. Another prominent research subject is the CAIPI algorithm for
itself. As the CAIPI algorithm can be considered as feedback-reliable data augmentation
procedure, it could be continuously adjusted and modified. Here, research subjects can be
instance selection, local explanation, or data augmentation methods. More sophisticated
methods than simple IoU are necessary to estimate the visual explanation quality more
accurately.

Further adjustments can be separated into three groups. First, the interface can be
evaluated in psychological studies. Second, the computational efficiency of XIML methods
can be increased by connecting them with online learning algorithms such as
\cite{Sahoo2018}. And third, the connection of inductive logic programming like in
\cite{Schmid2020} with human-in-the-loop ML procedures is a promising research area.
%
%
%
%
%
\bibliographystyle{splncs04}
\bibliography{references}

\begin{thebibliography}{10}
\providecommand{\url}[1]{\texttt{#1}}
\providecommand{\urlprefix}{URL }
\providecommand{\doi}[1]{https://doi.org/#1}

\bibitem{Bastani2017}
Bastani, O., Kim, C., Bastani, H.: Interpreting blackbox models via model
  extraction  (2017), \url{http://arxiv.org/abs/1705.08504}

\bibitem{Buslaev2020}
Buslaev, A., Iglovikov, V.I., Khvedchenya, E., Parinov, A., Druzhinin, M.,
  Kalinin, A.A.: Albumentations: Fast and flexible image augmentations. Inf.
  \textbf{11}(2), ~125 (2020). \doi{10.3390/info11020125}

\bibitem{Holzinger2016}
Holzinger, A.: Interactive machine learning for health informatics: when do we
  need the human-in-the-loop? Brain Informatics  \textbf{3}(2),  119--131
  (2016). \doi{10.1007/s40708-016-0042-6}

\bibitem{Hund2015}
Hund, M., Sturm, W., Schreck, T., Ullrich, T., Keim, D.A., Majnaric, L.,
  Holzinger, A.: Analysis of patient groups and immunization results based on
  subspace clustering. In: Guo, Y., Friston, K.J., Faisal, A.A., Hill, S.L.,
  Peng, H. (eds.) Brain Informatics and Health - 8th International Conference,
  {BIH} 2015, London, UK, August 30 - September 2, 2015. Proceedings. Lecture
  Notes in Computer Science, vol.~9250, pp. 358--368. Springer (2015).
  \doi{10.1007/978-3-319-23344-4\_35}

\bibitem{Kieseberg2015}
Kieseberg, P., Schantl, J., Fr{\"{u}}hwirt, P., Weippl, E.R., Holzinger, A.:
  Witnesses for the doctor in the loop. In: Guo, Y., Friston, K.J., Faisal,
  A.A., Hill, S.L., Peng, H. (eds.) Brain Informatics and Health - 8th
  International Conference, {BIH} 2015, London, UK, August 30 - September 2,
  2015. Proceedings. Lecture Notes in Computer Science, vol.~9250, pp.
  369--378. Springer (2015). \doi{10.1007/978-3-319-23344-4\_36}

\bibitem{Kingma2015}
Kingma, D.P., Ba, J.: Adam: {A} method for stochastic optimization. In: Bengio,
  Y., LeCun, Y. (eds.) 3rd International Conference on Learning
  Representations, {ICLR} 2015, San Diego, CA, USA, May 7-9, 2015, Conference
  Track Proceedings (2015), \url{http://arxiv.org/abs/1412.6980}

\bibitem{Kourou2015}
Kourou, K., Exarchos, T., Exarchos, K., Karamouzis, M., Fotiadis, D.: Machine
  learning applications in cancer prognosis and prediction. Computational and
  Structural Biotechnology Journal  \textbf{13} (11 2014).
  \doi{10.1016/j.csbj.2014.11.005}

\bibitem{Lozano2017}
Lozano, A.P.: {M}edical {MNIST} {C}lassification.
  \url{https://github.com/apolanco3225/Medical-MNIST-Classification} (2017)

\bibitem{Plumb2018}
Plumb, G., Molitor, D., Talwalkar, A.S.: Model agnostic supervised local
  explanations. In: Bengio, S., Wallach, H.M., Larochelle, H., Grauman, K.,
  Cesa{-}Bianchi, N., Garnett, R. (eds.) Advances in Neural Information
  Processing Systems 31: Annual Conference on Neural Information Processing
  Systems 2018, NeurIPS 2018, December 3-8, 2018, Montr{\'{e}}al, Canada. pp.
  2520--2529 (2018),
  \url{https://proceedings.neurips.cc/paper/2018/hash/b495ce63ede0f4efc9eec62cb947c162-Abstract.html}

\bibitem{Ribeiro2016}
Ribeiro, M.T., Singh, S., Guestrin, C.: "why should {I} trust you?": Explaining
  the predictions of any classifier. In: Krishnapuram, B., Shah, M., Smola,
  A.J., Aggarwal, C.C., Shen, D., Rastogi, R. (eds.) Proceedings of the 22nd
  {ACM} {SIGKDD} International Conference on Knowledge Discovery and Data
  Mining, San Francisco, CA, USA, August 13-17, 2016. pp. 1135--1144. {ACM}
  (2016). \doi{10.1145/2939672.2939778}

\bibitem{Rudin2019}
Rudin, C.: Stop explaining black box machine learning models for high stakes
  decisions and use interpretable models instead. Nat. Mach. Intell.
  \textbf{1}(5),  206--215 (2019). \doi{10.1038/s42256-019-0048-x}

\bibitem{Sahoo2018}
Sahoo, D., Pham, Q., Lu, J., Hoi, S.C.H.: Online deep learning: Learning deep
  neural networks on the fly. In: Lang, J. (ed.) Proceedings of the
  Twenty-Seventh International Joint Conference on Artificial Intelligence,
  {IJCAI} 2018, July 13-19, 2018, Stockholm, Sweden. pp. 2660--2666. ijcai.org
  (2018). \doi{10.24963/ijcai.2018/369}

\bibitem{Schmid2020}
Schmid, U., Finzel, B.: Mutual explanations for cooperative decision making in
  medicine. K{\"{u}}nstliche Intelligenz  \textbf{34}(2),  227--233 (2020),
  \url{https://doi.org/10.1007/s13218-020-00633-2}

\bibitem{Schramowski2020}
Schramowski, P., Stammer, W., Teso, S., Brugger, A., Herbert, F., Shao, X.,
  Luigs, H., Mahlein, A., Kersting, K.: Making deep neural networks right for
  the right scientific reasons by interacting with their explanations. Nat.
  Mach. Intell.  \textbf{2}(8),  476--486 (2020).
  \doi{10.1038/s42256-020-0212-3}

\bibitem{Settles2012}
Settles, B.: Active Learning. Synthesis Lectures on Artificial Intelligence and
  Machine Learning, Morgan {\&} Claypool Publishers (2012).
  \doi{10.2200/S00429ED1V01Y201207AIM018}

\bibitem{Teso2019}
Teso, S., Kersting, K.: Explanatory interactive machine learning. In: Conitzer,
  V., Hadfield, G.K., Vallor, S. (eds.) Proceedings of the 2019 {AAAI/ACM}
  Conference on AI, Ethics, and Society, {AIES} 2019, Honolulu, HI, USA,
  January 27-28, 2019. pp. 239--245. {ACM} (2019).
  \doi{10.1145/3306618.3314293}

\bibitem{Vedaldi2008}
Vedaldi, A., Soatto, S.: Quick shift and kernel methods for mode seeking. In:
  Forsyth, D.A., Torr, P.H.S., Zisserman, A. (eds.) Computer Vision - {ECCV}
  2008, 10th European Conference on Computer Vision, Marseille, France, October
  12-18, 2008, Proceedings, Part {IV}. Lecture Notes in Computer Science,
  vol.~5305, pp. 705--718. Springer (2008). \doi{10.1007/978-3-540-88693-8\_52}

\bibitem{MNIST2021}
Yang, J., Shi, R., Ni, B.: {MedMNIST} {C}lassification {D}ecathlon: {A}
  {L}ightweight {AutoML} {B}enchmark for {M}edical {I}mage {A}nalysis. In: IEEE
  18th International Symposium on Biomedical Imaging (ISBI). pp. 191--195
  (2021)

\bibitem{FashionMNIST}
{Zalando SE}: {F}ashion {MNIST}.
  \url{https://www.kaggle.com/zalando-research/fashionmnist} (2017)

\end{thebibliography}
\end{document}